%% file: main_wNames.tex
\documentclass[10pt,twocolumn,letterpaper]{article}

\usepackage{cvpr}              
\input{preamble}

\definecolor{cvprblue}{rgb}{0.21,0.49,0.74}
\usepackage[pagebackref,breaklinks,colorlinks,citecolor=cvprblue]{hyperref}

\usepackage{graphicx}
\usepackage{amsmath}
\usepackage{amssymb}
\usepackage{booktabs}
\usepackage[misc]{ifsym} 
\usepackage[accsupp]{axessibility}
\usepackage{verbatim}
\usepackage{multirow}
\usepackage{hhline}
\usepackage{booktabs}
\usepackage{xcolor}

\usepackage[capitalize]{cleveref}
\crefname{section}{Sec.}{Secs.}
\Crefname{section}{Section}{Sections}
\Crefname{table}{Table}{Tables}
\crefname{table}{Tab.}{Tabs.}

\makeatletter
\def\thanks#1{\protected@xdef\@thanks{\@thanks
        \protect\footnotetext{#1}}}
\makeatother

\title{Revisiting Single Image Reflection Removal In the Wild}

\begin{document}
\author{Yurui Zhu \textsuperscript{\rm 1, 2, $^{\star}$},
       Xueyang Fu \textsuperscript{\rm 1, $^{\textrm{\Letter}}$},
	Peng-Tao Jiang \textsuperscript{\rm 2}, \\
	Hao Zhang \textsuperscript{\rm 2},
	  Qibin Sun \textsuperscript{\rm 1},
   Jinwei Chen  \textsuperscript{\rm 2},
   Zheng-Jun Zha  \textsuperscript{\rm 1},
  Bo Li \textsuperscript{\rm 2, $^{\textrm{\Letter}}$}  \\
\textsuperscript{\rm 1} University of Science and Technology of China \quad 
 \textsuperscript{\rm 2} vivo Mobile Communication Co., Ltd  \\ 
{\tt\small zyr@mail.ustc.edu.cn, xyfu@ustc.edu.cn, libra@vivo.com 
}
}
\thanks{ 
\hspace{-6mm}  ${\star}$ : This work was done during his internship at vivo Mobile Communication Co., Ltd. \\
 ${\textrm{\Letter}}$ : Corresponding authors.  }

\maketitle

\begin{abstract}

This research focuses on the issue of single-image reflection removal (SIRR) in real-world conditions, examining it from two angles:  the collection pipeline of real reflection pairs and the perception of real reflection locations.
We devise an advanced reflection collection pipeline that is highly adaptable to a wide range of real-world reflection scenarios and incurs reduced costs in collecting large-scale aligned reflection pairs.
In the process, we develop a large-scale, high-quality reflection dataset named Reflection Removal in the Wild (RRW). RRW contains over 14,950 high-resolution real-world reflection pairs, a dataset forty-five times larger than its predecessors. 
Regarding perception of reflection locations, we identify that numerous virtual reflection objects visible in reflection images are not present in the corresponding ground-truth images. 
This observation, drawn from the aligned pairs, leads us to conceive the Maximum Reflection Filter (MaxRF). The MaxRF could accurately and explicitly characterize reflection locations from pairs of images. Building upon this, we design a reflection location-aware cascaded framework, specifically tailored for SIRR. 
Powered by these innovative techniques, our solution achieves superior performance than current leading methods across multiple real-world benchmarks. Codes and datasets will be publicly available.  
\end{abstract}

\input{Sec1_intro_re}

\input{Sec2_Related_work}

\input{Sec2.5_Revisit}

\input{Sec3_method}

\input{Sec4_experiment}

\input{Sec5_Conclusion}

{
    \small
    \bibliographystyle{ieeenat_fullname}
    \bibliography{CVPR2024-reflection/egbib}
}
\end{document}

%% file: preamble.tex
\usepackage[dvipsnames]{xcolor}


%% file: Sec1_intro_re.tex
\section{Introduction}


\begin{figure}[t]
\captionsetup{type=figure}
\centering
\resizebox{\linewidth}{!}{
\includegraphics[width=1.0\textwidth]{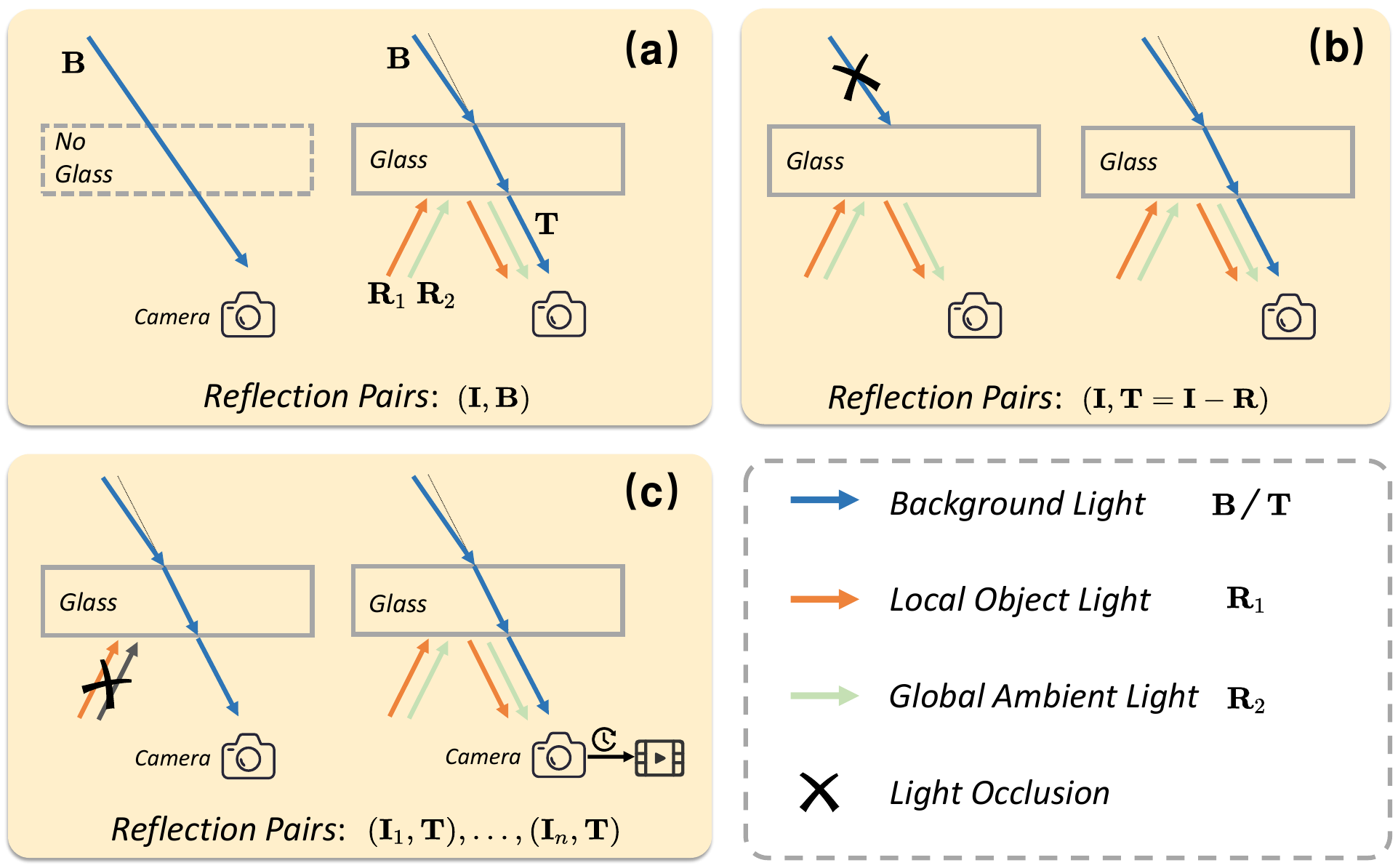}}
\vspace{-12pt}
\caption{Simplified illustrations of existing pipelines for collecting real reflection pairs (\textbf{I}: reflection image). (a) the pipeline from~\cite{zhang2018single,li2020single,wan2017benchmarking}, which may lead to the misalignment between $\textbf{B}$ and $\textbf{T}$ due to the glass refraction. (b) the pipeline from~\cite{lei2020polarized,lei2022categorized}. Their pipeline is only suitable for collection in the RAW data format, and the obtained  $\textbf{T}$ may exist reflection remnant artifacts. (c) Ours, which avoids interference from glass refraction or artifacts and does not require additional data format specifications. Moreover, the video-based capture system would help to reduce the difficulty and effort involved in large-scale data acquisition.}
\vspace{-5pt}
\label{fig: pipelines}
\end{figure}

In photographic environments involving reflective materials, such as glass, the inadvertent emergence of reflections is a common challenge. These underside reflections not only diminish the aesthetic quality of the captured images but also impede the accuracy of follow-up computer vision tasks~\cite{sinha2012image, liu2020reflection,qiu2023looking}. Consequently, devising effective reflection removal algorithms is important and meaningful.

\begin{figure}[t]
\captionsetup{type=figure}
\centering
\resizebox{\linewidth}{!}{
\includegraphics[width=1.0\textwidth]{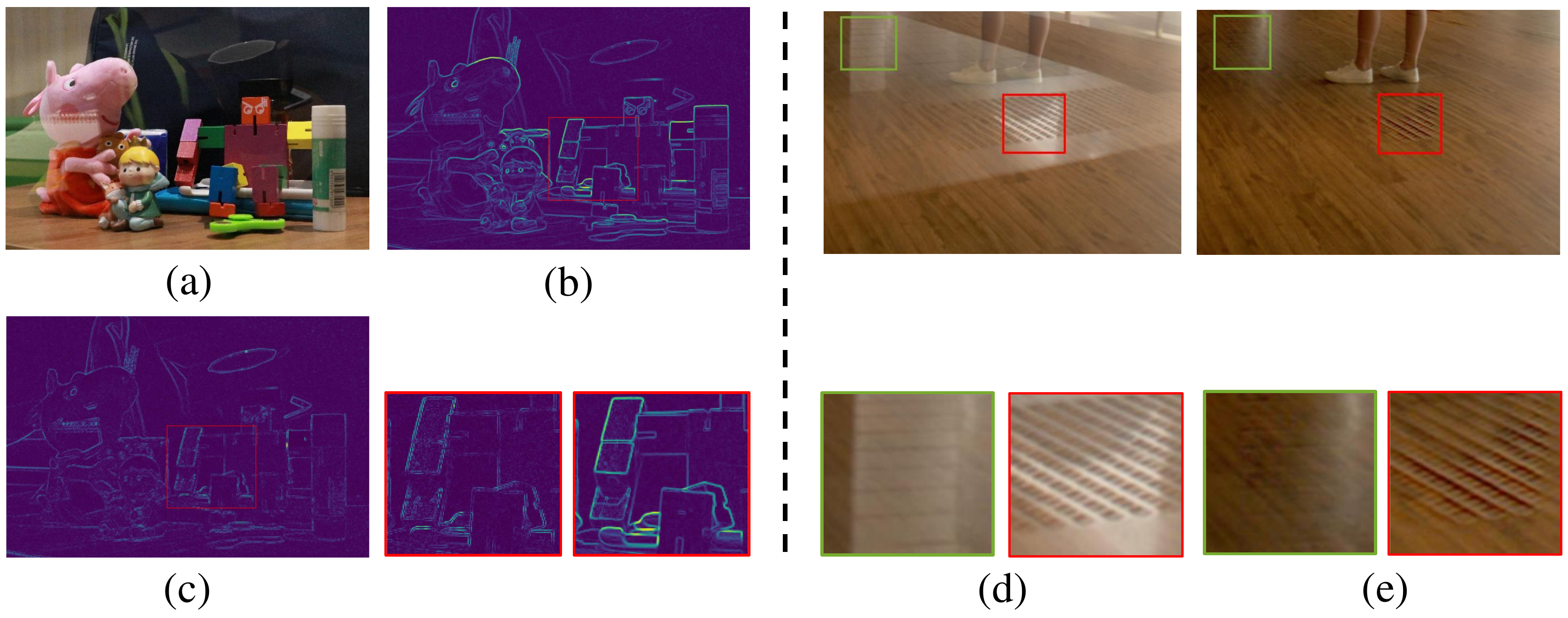}}
\vspace{-18pt}
\caption {Analysis of collected images from the previous collection pipelines.
(a) Reflection image from the acquisition pipeline~\cite{zhang2018single,li2020single,wan2017benchmarking}; (b) Gradient of (a); (c) Gradient difference map between the reflection pair, where the 'double edge' in the gradient difference map indicates misalignment due to glass refraction. (d) and (e) are the reflection pair from the acquisition pipeline~\cite{lei2020polarized}, where minor reflection remnants could be found in the corresponding transmission images.
(\textbf{Best viewed on screen.)}} 
\vspace{-5pt}
\label{fig: drawbacks}
\end{figure}

Deep learning-based SIRR methods have recently exhibited encouraging results. It's well acknowledged that an ample supply of high-quality data is essential for these data-driven methods. 
Accordingly, a range of datasets has been developed to support research in SIRR. Nevertheless, our thorough examination of the collection pipelines corresponding to these datasets reveals consistently overlooked issues in Figure~\ref{fig: pipelines}.
For instance, pioneering studies ~\cite{zhang2018single,wei2019single,li2020single} acquire reflection-free images by manually removing the glass. However, this technique invariably induces spatial pixel misalignment due to the refraction triggered by the glass. Lei \etal ~\cite{lei2020polarized, lei2022categorized} exploit the linear reflection formation in raw space to extract the transmission layer with the subtraction operation. However, as depicted in Figure~\ref{fig: drawbacks}, we noticed that their method might leave minor reflection remnants in the corresponding transmission images. Furthermore, it's worth noting that the effort and time necessary to expand the dataset size based on previous pipelines can be quite costly. Therefore, the costs associated with prior data collection pipelines, along with issues like misalignment or artifacts, cause a shortage of high-quality, large-scale real pairs for training deep models.

In this study, we revisit the reflection physical formulation process. This subsequently leads to the development of an innovative pipeline for the collection of real reflection pairs.
To be specific, the essence of reflection disturbances lies in the fact that the reflected lights pass through the surface of the reflective material and are captured by camera devices, thereby disturbing the clarity of the transmission layer. 
Hence, we could directly obtain reflection-free images by obstructing the reflective light, thereby enabling the acquisition of pairs of images with and without reflections.
As illustrated in Figure~\ref{fig: pipelines}, our pipeline captures the transmission layer without the removal of glasses, utilizing non-transmissive and non-reflective black cloth to block reflection lights. 
Subsequently, the light-blocking materials are removed, facilitating the capture of images with reflection distortions. It is noteworthy that the whole process is implemented in video mode, reducing the difficulty and efforts involved in large-scale data acquisition. During the collection phase, we could dynamically manipulate reflective contents, thereby enhancing the diversity of reflection distortions. Evidently, compared to other pipelines, our pipeline enables us to more easily acquire the aligned and extensive real reflection pairs for training deep networks.

Upon further revisiting the reflection imaging process, we identify that the constituents of reflections can be categorized into two parts: global reflections induced by ambient light and local reflections caused by specific objects. 
The latter, local reflections, are typically more challenging, prompting many innovative solutions ~\cite{wan2018region, dong2021location} to address them. 
For example, Wan \etal~\cite{wan2018region} assume that the gradients of reflections are generally small and obtain the reflection-dominated regions via the threshold algorithm. Indeed, such an approach is not suitable for scenarios with strong reflections.  Dong \etal ~\cite{dong2021location} utilize the linear composition loss to implicitly infer the reflection confidence maps. Yet, their linear assumption often falls short in describing the complex real-world scenes. 
Such location cues are known to effectively mitigate reflection disturbances, as demonstrated by studies~\cite{wan2017benchmarking, dong2021location, wan2018region}.
However, these techniques rely on prior assumptions to indirectly obtain the desired results, which often come with certain limitations.

Currently, it is rarely investigated the explicit representation to locate the reflection regions without specific assumptions. Nonetheless, in this study, we highlight that if the collection pipelines enable collecting aligned reflection pairs, then directly utilizing these aligned paired images can effectively characterize the locations of these local reflections.
Specifically, we observe the fact that the reflection layer encompasses textures of many virtual objects, which are absent in the corresponding ground-truth images. 
Building upon this observation, we devise the maximum reflection filter (MaxRF) that could explicitly present the reflection locations. Therefore, this paper proposes a divide-and-conquer framework tailored for SIRR, including reflection detection and removal. Within our frameworks, we distinguish the local reflection regions based on representations derived from MaxRF. Subsequently, these location cues are integrated into the second stage of our framework, significantly enhancing the performance of removing reflections. 
Finally, experimental results also demonstrate the effectiveness and the superiority of our proposed solution. 
Contributions of this paper could be summarized as:
\begin{itemize}
  \item We propose a new pipeline for the collection of real reflection pairs. notable for its adaptability to a wide range of reflection scenarios and its independence from data format constraints. This pipeline also offers a more cost-effective manner of acquiring reflection datasets.
  
  \item We present a large-scale high-quality paired reflection dataset, Reflection Removal in the Wild (RRW). To the best knowledge, RRW is the largest paired reflection dataset, comprising 14952 pairs of high-resolution images captured across diverse real-world reflection scenes.

  \item We propose the maximum reflection filter (MaxRF), which enables obtaining the explicit location representation to characterize reflection regions. 

  \item We develop a cascaded network for SIRR, which involves the reflection detection and removal network, \ie, first learning to estimate reflection locations with MaxRF and then removing reflections with location guidance. Comprehensive experiments indicate the superiority of the proposed innovations. 
\end{itemize}


%

%% file: Sec2_Related_work.tex
\section{Related work}

Over recent decades, numerous innovative methods have been proposed to tackle the issue of image reflection removal. Some approaches usually require additional inputs, such as multi-frames~\cite{li2020improved,ahmed2021user,niklaus2021learned}, polarization~\cite{schechner2000polarization, lei2020polarized,nayar1997separation}, and flash-only prior~\cite{lei2021robust,lei2023robust}. In this paper, our primary focus lies in the field of single-image reflection removal.

\textbf{Traditional methods}. 
Early methods ~\cite{li2014single, shih2015reflection, xue2015computational, lei2021robust,zheng2021single, levin2007user,arvanitopoulos2017single,guo2014robust} utilize various image priors to eliminate reflection degradation. For example,  Li \etal ~\cite{li2014single} devise a relative smoothness prior, postulating that the reflection contents are intrinsically blurry, consequently penalizing these larger gradients.  Shih\etal ~\cite{shih2015reflection} propose to automatically suppress reflection by leveraging "ghosting" cues from double reflections on thick glasses and employing a Gaussian Mixture Model for regularization. In ~\cite{levin2007user}, the user annotations are used to guide layer separation between the transmission and reflection layers.
Nikolaos \etal ~\cite{arvanitopoulos2017single} impose the laplacian data fidelity term and gradient sparsity as optimization objectives. Despite producing decent results, traditional methods often rely on assumptions and tend to have slower processing speeds. 

\textbf{Learning-based methods}.
With the development of deep-learning techniques, learning-based SIRR methods~\cite{wan2019corrn, liu2019semantic, ma2019learning,wan2018crrn,prasad2021v,wan2020reflection,hu2023single} also achieve great performance gains in the diverse reflection scenes and dominate this field. Concretely, CEILNet ~\cite{fan2017generic} adopts a two-stage network approach, which first estimates the edge map and subsequently reconstructs the transmission layer.  ERRNet~\cite{wei2019single} attempts to leverage high-level contextual features to mitigate uncertainty in these regions with prominent reflections and introduce the misaligned real-world pair images. Yu \etal~\cite{li2023two} and BDNet ~\cite{yang2018seeing} both incorporate the reflection layer to guide the restoration of the transmission layer. Song \etal~\cite{dong2021location} further investigates the robustness of SIRR networks against adversarial attacks. In addition, previous methods~\cite{hu2021trash, hu2023single} have explored the complementary mechanism and developed the dual-stream frameworks to achieve reflection separation. Moreover, LANet~\cite{dong2021location} proposes a location-aware solution with a recurrent network to remove reflections in single images, improving results by emphasizing strong reflection boundaries with Laplacian features. Unlike LANet using the implicit manner to estimate the reflections, our proposed method perceives the reflection location with the explicit representation.

Various methods are also devoted to addressing the insufficiency of real-world training data, categorizing them into three primary avenues.  The first direction is to model reflections that are more consistent with real-world scenarios. For example, Kim\etal ~\cite{kim2020single} utilize the physically-based rendering to generate various reflection image pairs for training.  Wen \etal  ~\cite{wen2019single} propose to exploit the non-linearity capability of deep neural networks to simulate the real-world physical reflection imaging process. However, these synthetic reflection images are still far from the real-world physical formulation, which may bring the performance drop under real reflection scenes.  The second approach utilizes unsupervised and weakly supervised algorithms to alleviate the demand for large-scale paired training data. ~\cite{rahmanikhezri2022unsupervised} proposed an unsupervised SIRR method by optimizing the two deep network parameters to separate the target image into exclusive transmission and reflection layers. However, compared to supervised learning techniques, the performance of these methods~\cite{rosh2023deep, rahmanikhezri2022unsupervised} still exhibits considerable room for further advancement.

The third direction involves collecting paired reflections from the real world. For example, methods ~\cite{zhang2018single,li2020single,wan2017benchmarking} capture pair samples where images with the manual arrangement of glass represent reflection images, while those without the glass serve as reflection-free images. Moreover, ~\cite{lei2020polarized, lei2022categorized} observe the linear formulation of the reflection imaging process is held on raw space, and they attempt to obtain the transmission layer by subtraction operation. However, in this paper, compared with previous ones, our pipeline could avoid the pixel misalignment introduced by glass refraction and does not make assumptions regarding the data format. Moreover, our pipeline is more applicable for diverse reflective scenes and enables the acquisition of large-scale reflection image pairs at a lower cost.



%% file: Sec2.5_Revisit.tex
\begin{figure}[t]
\captionsetup{type=figure}
\centering
\resizebox{\linewidth}{!}{
\includegraphics[width=1.0\textwidth]{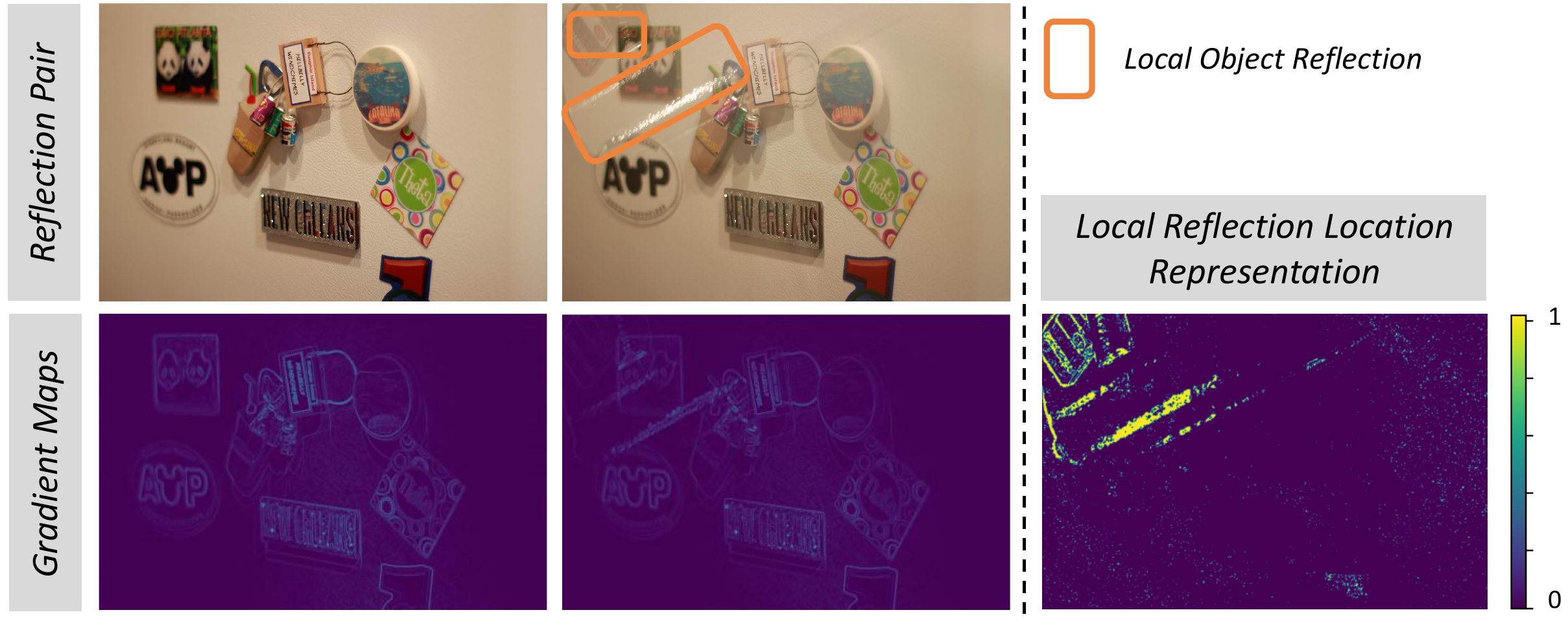}}
\vspace{-10pt}
\caption {Visualization of the local reflection location presentation via our proposed Maximum Reflection Filter (MaxRF). 
First column: non-reflection images. Second column: reflection images
The global ambient reflection causes attenuation of the color and contrast, subsequently diminishing the gradient intensity of the original transmission contents. Hence, applying MaxRF predominantly could well highlight local reflection locations.
(\textbf{Best viewed on screen.}) 
}
\vspace{-5pt}
\label{fig:local_location}
\end{figure}

\section{Revisit Reflection Physical Formulation}


We revisit the physical formulation underlying the occurrence of reflections, specifically using reflection scenarios involving glasses reflection as examples. Within the reflection-contaminated image, the physical lights are a mixture of reflection and transmission lights. In this section, we further analyze these two components and clarify ambiguities of SIRR identified in previous studies.

For the former, we first define the light originating from behind the glasses as the background $\textbf{B}$, and the light that passes through the glass as the transmission $\textbf{T}$. As cameras shoot these objects behind the glasses, the background light undergoes refraction and absorption~\cite{zheng2021single}, subsequently transforming into the transmission light.
This results in a distinct difference between the background $\textbf{B}$  and the transmissions $\textbf{T}$, as depicted in Figure~\ref{fig: pipelines}. Notably, the inherent refraction property of glass indicates that earlier data acquisition pipelines\cite{zhang2018single,wei2019single,li2020single} inherently led to pixel misalignment, making it impossible to obtain perfectly aligned image pairs, especially thick glasses. Moreover, prevailing studies~\cite{yan2023single, yano2010image} suggest that the light transmitted through glass substantially surpasses the absorbed part, indicating the absorption effects can be roughly neglected. Therefore, according to the above analysis, we argue that SIRR should focus on removing the illumination disturbances from the camera's side and obtaining clean transmission $\textbf{T}$, which is also consistent with ~\cite{lei2020polarized}.

Besides, reflections commonly manifest when a camera captures illumination rays reflected off surfaces within its field of view. We noted that physical aspects of reflections can be broadly categorized into two distinct components. As shown in the first row of Figure~\ref{fig:local_location}, the first involves global reflection resulting from ambient light, causing attenuation of the color and contrast in the captured images. The second pertains to the virtual contents formed after the reflection from objects on the camera side, which typically occupies a part of the whole image. Such local reflections typically result in occlusions or overlapping with the transmission contents. Previous methods attempt to identify the spatial location of these local reflections either through the indirect paradigm, \eg, gradient prior~\cite{wan2018region} or via implicit constraints~\cite{dong2021location,li2020single}. In contrast, we propose the maximum reflection filter(MaxRF) to directly acquire reflection locations from reflection pairs. Subsequently, we employ neural networks to learn and distinguish such local reflection regions, as discussed in Sec.~\ref{subsec: location}. 




%% file: Sec3_method.tex
\section{Method}


\subsection{Explicit Reflection Location Perception}
\label{subsec: location}

The virtual image formed by object reflections often occupies only a portion of the image, and it is typically the more challenging reflection component to handle. In this paper, we devise an explicit representation to characterize these reflection locations. 

Due to the global ambient light reflection, the difference between the reflection pairs cannot directly obtain the local reflection location information. However, in fact, regardless of the strength of these local reflections, it often results in the presence of texture details in the reflection-contaminated image that is absent in the transmission layer. Hence, we propose the Maximum Reflection Filter (MaxRF) to identify these reflection regions. To elaborate, MaxRF involves two key steps. Firstly, we apply the Sobel operator~\cite{gonzales1987digital} to compute gradient maps for both the reflection and reflection-free images. Secondly, we perform maximum comparisons in the corresponding gradient domain. It's worth noting that due to the influence of global ambient light reflection, the intensity of gradients associated with the original transmissions is also reduced compared to their initial strength. As a result, when we apply the MaxRF, what remains mainly are the gradients indicative of the local reflection contents.
Finally, we could employ the reflection image pairs to explicitly obtain the local reflection locations, denoted as $\textbf{M}_{local}$, which could be expressed as:

\begin{equation}
\label{eqn:location}
\begin{aligned}
   G_{I}, G_{T}  &= Grad(\textbf{I}), Grad(\textbf{T}), \\
   \textbf{M}_{local}^{(i,j)} &= 
\begin{cases} 
1 & \text{if } G_{I}^{(i,j)} > G_{T}^{(i,j)}, \\
0 & \text{otherwise},
\end{cases} 
\end{aligned}
\end{equation}
where $\textbf{I}$ and $\textbf{T}$ indicate the reflection-contaminated image and transmission image, respectively; $Grad(\cdot )$ indicates the Sobel gradient operator; and $\textbf{M}_{local}^{(i,j)} = 1$ indicates the presence of the local reflection at the spatial position $(i,j)$. Moreover,
we also visualize the local reflection location presentation via our proposed (MaxRF) in Figure ~\ref{fig:local_location}.

\begin{figure}[t]
\captionsetup{type=figure}
\centering
\resizebox{\linewidth}{!}{
\includegraphics[width=1.0\textwidth]{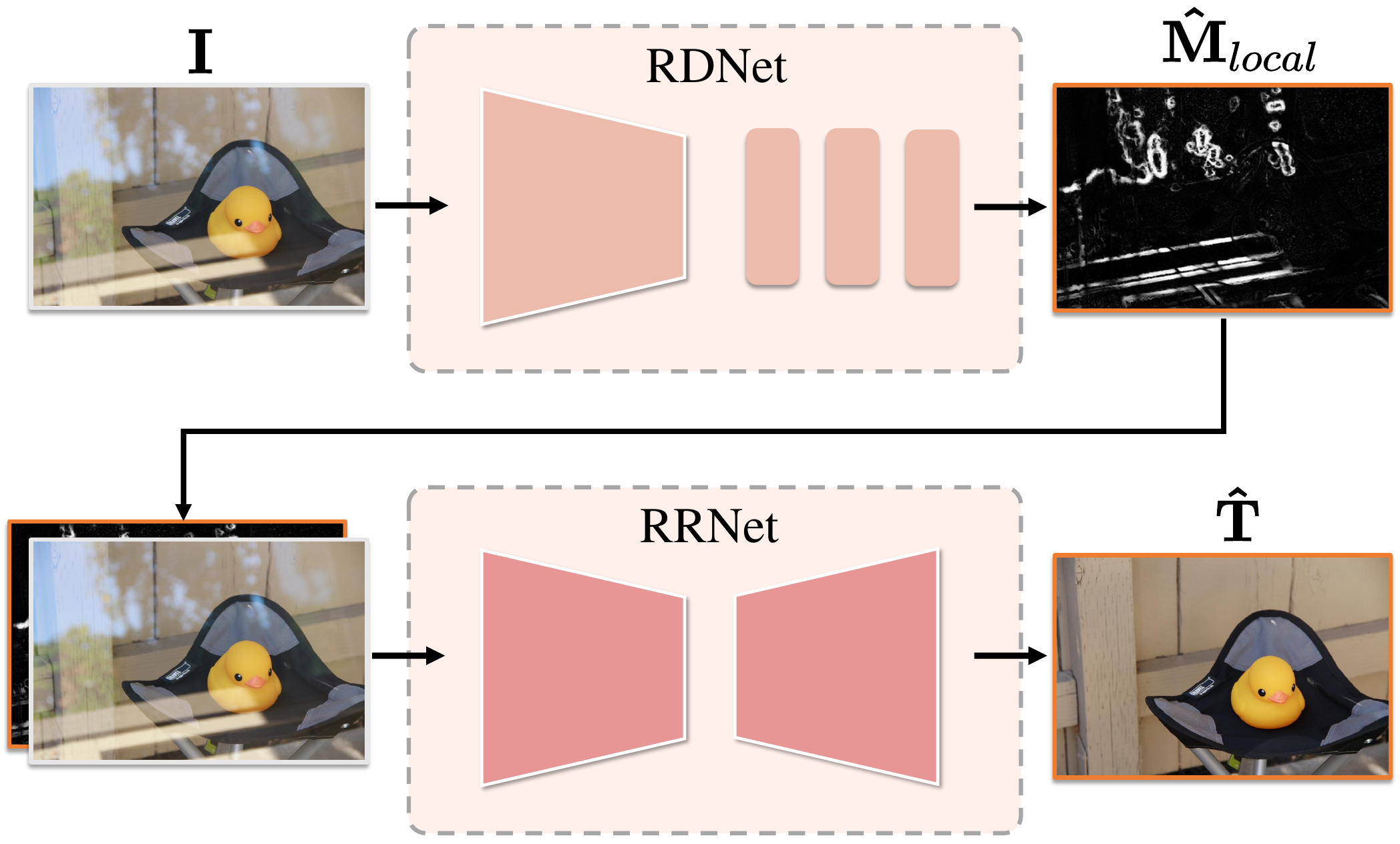}}
    \vspace{-13pt}
\caption {Simplified illustration of our proposed framework, including the RDNet (Reflection Detection Network) and RRNet (Reflection Removal Network).}
\vspace{-5pt}
\label{fig: framework}
\end{figure}

\subsection{Network Architecture}

In this section, we further elaborate the proposed cascaded network architecture tailored for SIRR. As shown in figure~\ref{fig: framework}, this structure comprises two primary components: the reflection detection network (RDNet) and the reflection removal network (RRNet). 
For RDNet, we employ the pre-trained network backbone~\cite{tan2019efficientnet} in conjunction with several residual blocks~\cite{he2016deep} and interpolation operation to estimate $\mathbf{M}_{local}$. For RRNet, we utilize the estimated $\mathbf{\hat{M}}_{local}$ to guide the subsequent reflection removal. Concretely, we adopt the widely used restoration backbone~\cite{chen2022simple} as our baseline structure for reflection removal. The estimated $\mathbf{\hat{M}}_{local}$ and the input reflection image $\mathbf{I}$ are concatenated and subsequently fed into RRNet. The whole process could be formulated as:
\begin{align}
\label{eqn:dnet}
   \mathbf{\hat{M}}_{local} & = \text{RDNet}(\mathbf{I}), \\
\label{eqn:rnet}
   \hat{\mathbf{T}} & = \text{RRNet}(Concat([\mathbf{I},  \mathbf{\hat{M}}_{local} ])).
\end{align}

Based on Eqn.~\ref{eqn:location}, we can directly utilize the reflection image pair to derive the explicit representation for local reflection regions. Consequently, we adopt the supervised learning manner to estimate $\mathbf{\hat{M}}_{local}$. The loss function for RDNet is defined as follows:
\begin{equation}
\label{eqn:loss_DNet}
\mathcal{L}_{DNet} = \left \|  \mathbf{M}_{local}   -  \mathbf{\hat{M}}_{local}  \right \| _{1}  + \gamma_{1} * TVLoss(\mathbf{\hat{M}}_{local}),
\end{equation}
where $  \gamma_{1} $ represents the balancing weight for the $TVLoss$~\cite{rudin1992nonlinear}, adopted to smooth the estimated results and mitigate artifacts. Moreover, for RRNet, we employ the content loss and perceptual loss as defined in~\cite{zhang2018single,li2020single,hu2023single}, which are expressed as:
\begin{equation}
\label{eqn:loss_RNet}
\mathcal{L}_{RNet} = \left \|  \mathbf{T}  -  \hat{\mathbf{T}} \right \| _{1}  +  \gamma_{1} * \left \| VGG(\mathbf{T})  -  VGG(\hat{\mathbf{T}}) \right \| _{1},
\end{equation}
where $ \gamma_{2} $ indicates the balanced weight; $VGG(\cdot)$ indicates hierarchical features extracted by the four layers $conv1\_2$, $conv2\_2$, $conv3\_2$, and $conv4\_2$ of the VGG19 ~\cite{simonyan2014very}.

\subsection{Dataset Collection Pipeline}

As illustrated in Figure~\ref{fig: pipelines}, we compare our proposed pipeline with those previously presented in~\cite{zhang2018single,li2020single,lei2020polarized}. Figure~\ref{fig: pipelines}(a) employs images captured both before and after the manual placement of glass, which facilitates capturing both reflection-containing images and their background counterparts. Nonetheless, their approach fails to consider the refraction effect and color of the glasses. Consequently, when the glasses are thick or colored, it can lead to noticeable pixel misalignment or color distortion. Meanwhile, following this pipeline, it is impractical to capture reflection image pairs by involving the common glass scenes in daily life, restricting its broad applicability. Moreover, such misaligned pairs can even pose challenges for network training~\cite{lei2020polarized}. On the other hand, both Figure~\ref{fig: pipelines}(b) and ours recognize the refraction effect of glass. However, the second pipeline is based on the observation of the linear physical composition held on raw data. They initially capture the reflection-contaminated image $\textbf{I}$ and the reflection layer $\textbf{R}$ in raw format, and then produce the transmission layer through $\textbf{T} = \textbf{I} - \textbf{R}$. However, empirically, we notice that the obtained transmission layers frequently retain subtle reflection remnants.

In contrast to the previous, our pipeline is applicable to a wide range of reflection scenarios (\eg, various glass scenes) and relaxes the requirement of the data format. Specifically, we commence the collection process by employing a black velvet cloth to block reflective lights on the camera side, ensuring the acquisition of the clean transmission image, denoted as $\textbf{T}$. Once the cloth is removed, we can then acquire images with reflection disturbances. More importantly, our pipeline is implemented in video mode. During this phase,  we further actively modulate the contents (\eg, ) on the reflective side to diversify the reflective scenes. These modulation operations include blocking reflective lights and adjusting or introducing reflective objects, among others.  Therefore, our proposed pipeline also facilitates the scaling up of real-world reflection training datasets at a lower cost.

\begin{table}[t]
\caption{Summary of existing real reflection datasets.}
\vspace{-10pt}
\label{tab:datasets}
\resizebox{\linewidth}{!}{
\begin{tabular}{lcccc}
\toprule
\multicolumn{1}{c}{Dataset} & Year & Usage & \begin{tabular}[c]{@{}c@{}}Pairs\\ Number\end{tabular} & \begin{tabular}[c]{@{}c@{}}Average\\ Resolution\end{tabular} \\ \hline
$SIR^2$ & 2017 & Test & 454 & 540 $\times$ 400 \\
$Real$ & 2018 & Train / Test & 89/20 & 1152 $\times$ 930 \\
$Nature$ & 2020 & Train / Test & 200/20 & 598 $\times$ 398 \\
$RRW$(Ours) & 2023 & Train & 14952 & 2580 $\times$ 1460 \\ \bottomrule
\end{tabular}
}
\vspace{-15pt}
\end{table}


Our proposed dataset is primarily captured using two camera devices: the Apple iPhone 13 and a Digital Single-Lens Reflex (DSLR) Canon EOS 200DII. In total, we have collected nearly 150 video clips and sampled 14952 pairs of reflection images. To ensure alignment between $\textbf{M}$ and $\textbf{T}$, the tripod and the remote control shutter are used for camera stabilization.
Moreover, we provide comparisons with other real-world datasets in Table~\ref{tab:datasets} and showcase some reflection examples in Figure~\ref{fig: examples}.  

%% file: Sec4_experiment.tex
\begin{figure*}[t]
\captionsetup{type=figure}
\centering
\resizebox{\linewidth}{!}{
\includegraphics[width=1.0\textwidth]{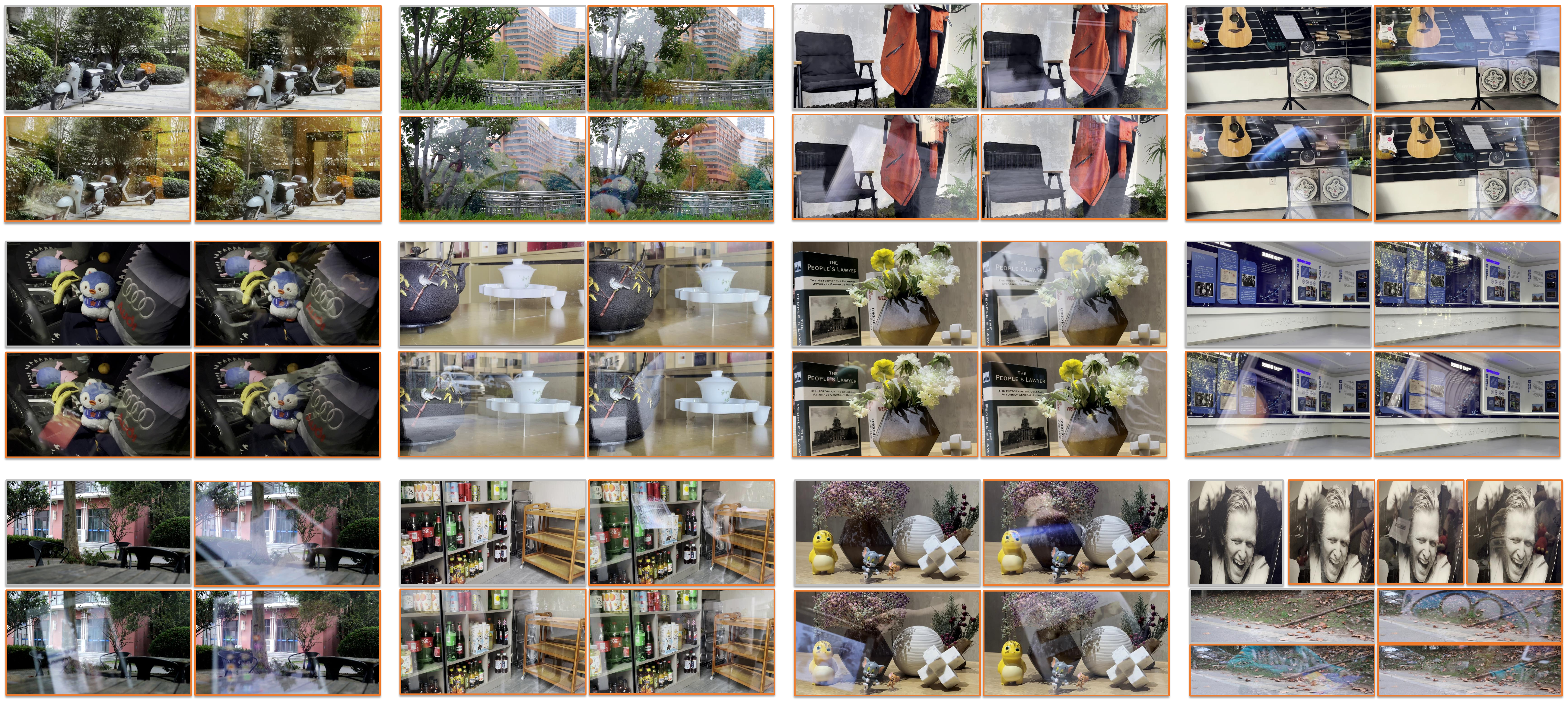}}
    \vspace{-5pt}
\caption {Some reflection pairs of the proposed RRW dataset. Our pipeline is a video mode capture system, where each non-reflection image (\textcolor{gray}{gray} boxes) corresponds to multiple real reflection images (\textcolor{orange}{orange} boxes), encompassing a variety of reflective surfaces such as building glasses, car glass windows, display glass, framing glass, and self-prepared glass. The diversity demonstrates the pipeline's applicability across various reflection scenarios. Best viewed on screen. }
\label{fig: examples}
\end{figure*}

\begin{figure*}[t]
\captionsetup{type=figure}
\centering
\resizebox{\linewidth}{!}{
\includegraphics[width=1.0\textwidth]{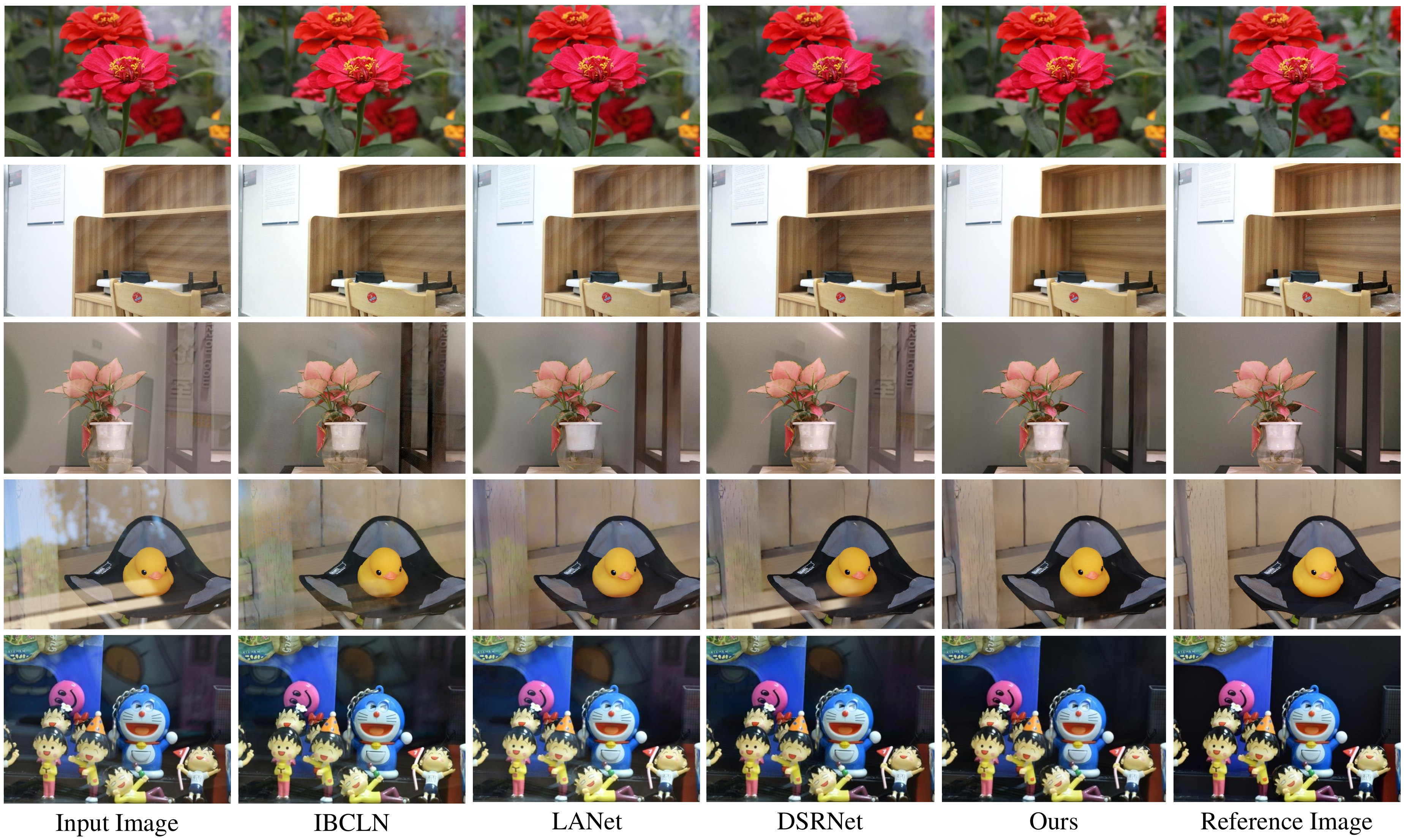}}
    \vspace{-5pt}
\caption {Visual comparisons between our method and previous methods. Unless otherwise specified, all reflection images in this paper are from real-world reflection scenes. More visual results are available in our supplemental material.}
\vspace{-5pt}
\label{fig: comparison1}
\end{figure*}

\input{tabs/performance_comparisons}

\section{Experiments}

\subsection{Implementation details}
Our framework is implemented with PyTorch platform on a PC with  NVIDIA GeForce GTX 1080Ti. At the training phase, the network is trained by Adam~\cite{kingma2015adam} optimizer with an initial learning rate of 0.0006, which changes based on Cosine Annealing scheme~\cite{loshchilov2016sgdr}. The batch size is set to 4, and the 320 $\times$ 320 patches are randomly cropped from the image at each training iteration. 
 The hyperparameters in Eqns.~\ref{eqn:loss_DNet}~\ref{eqn:loss_RNet} are empirically set as $\gamma_{1} = 0.00005$, $\gamma_{2} = 0.02$. 

\subsection{Dataset and Evaluation Metrics}

During the training phase, we enhance the training dataset by integrating data used in previous methods~\cite{li2020single,hu2021trash,hu2023single,dong2021location}, with additional data collected RRW dataset, providing a more comprehensive training database. For the testing dataset, following previous methods, we evaluate the performance of our model by applying one pre-trained reflection removal model across three real-world reflection benchmarks: $Real$, $Nature$, and $SIR^{2}$. These three benchmarks, developed through different works, comprehensively cover a variety of real-world reflection scenarios, which are widely used for showcasing the performance of models in real-world reflection scenarios. However, among these three datasets $Real$, $Nature$, and $SIR^{2}$, the first two include both training and test datasets, while the latter inherently serves as the test dataset.

Besides, we employ the peak signal-to-noise ratio (PSNR) and structural similarity (SSIM) \cite{wang2004image} as the evaluation metrics. These are calculated in the RGB color space, where higher values denote superior performance.

\subsection{Comparison to State-of-the-arts}
To evaluate the reflection removal performance, we compare our proposed method with 11 SIRR methods, including BDN ~\cite{yang2018seeing}, FRS ~\cite{yang2019fast}, Zhang \etal~\cite{johnson2016perceptual}, ERRNet~\cite{wei2019single}, RMNet~\cite{wen2019single}, 
Kim \etal~\cite{kim2020single},
IBCLN~\cite{li2020single}, YTMT~\cite{hu2021trash}, LANet~\cite{dong2021location},            PNACR~\cite{wang2023personalized}, and DSRNet~\cite{hu2023single}. For fair comparisons, we directly employ the pre-trained weights publically provided by their authors. Following~\cite{wei2019single, hu2023single}, the comparison experiments are also performed under the same settings, such as using the same reflection inputs and the same performance evaluation codes.   Additionally, for methods~\cite{johnson2016perceptual, wei2019single}, additional finetuning is implemented. This was due to the fact that these methods were developed before the introduction of the $Nature$ dataset, and as such, their initial training dataset does not encompass the training data of the $Nature$ dataset. Note that we do not perform finetuning on RMNet~\cite{wen2019single}, due to their method relays on additional alpha blending masks from SynNet~\cite{wen2019single}.
Hence, apart from the differences in the design of the algorithm compared to other methods, our solution further utilizes the collected real-world dataset RRW, to construct a more comprehensive training dataset.  

The quantitative comparison results are reported in Table~\ref{tab:performance}. Our proposed method obviously achieves the best PSNR scores across all real benchmarks, which effectively demonstrates superior performance against the previous SOTA methods. Specifically, our approach is 2.03dB higher than the second-best method~\cite{wang2023personalized} on the PNSR metric. The last column reports that our method also achieves the best average PSNR and SSIM scores. This verifies the powerful generalization capacity on the various real reflection cases.

Figure~\ref{fig: comparison1} further provides visual comparisons of reflection removal results from three SOTA methods and ours. These real reflection inputs all from $Real$, $Nature$, and $SIR^{2}$ dataset. For example, in the third row of input images, global ambient reflections and local object reflections are both present. IBCLN~\cite{li2020single} and LANet~\cite{dong2021location}, although significantly mitigating the global color and contrast distortions induced by global reflections, still exhibit residual local reflections. 
DSRNet~\cite{hu2023single}, while successful in diminishing certain local reflections, exhibits a noticeable disparity in color and contrast compared to the reference image. We observe in the fourth row that other methods, when dealing with stronger reflections, tend to introduce artifacts such as color biases. In contrast, our approach effectively eliminates both types of reflections and retains high-frequency transmission details in the estimated results.

\subsection{Ablation study}

\begin{figure}[t]
\captionsetup{type=figure}
\centering
\resizebox{\linewidth}{!}{
\includegraphics[width=1.0\textwidth]{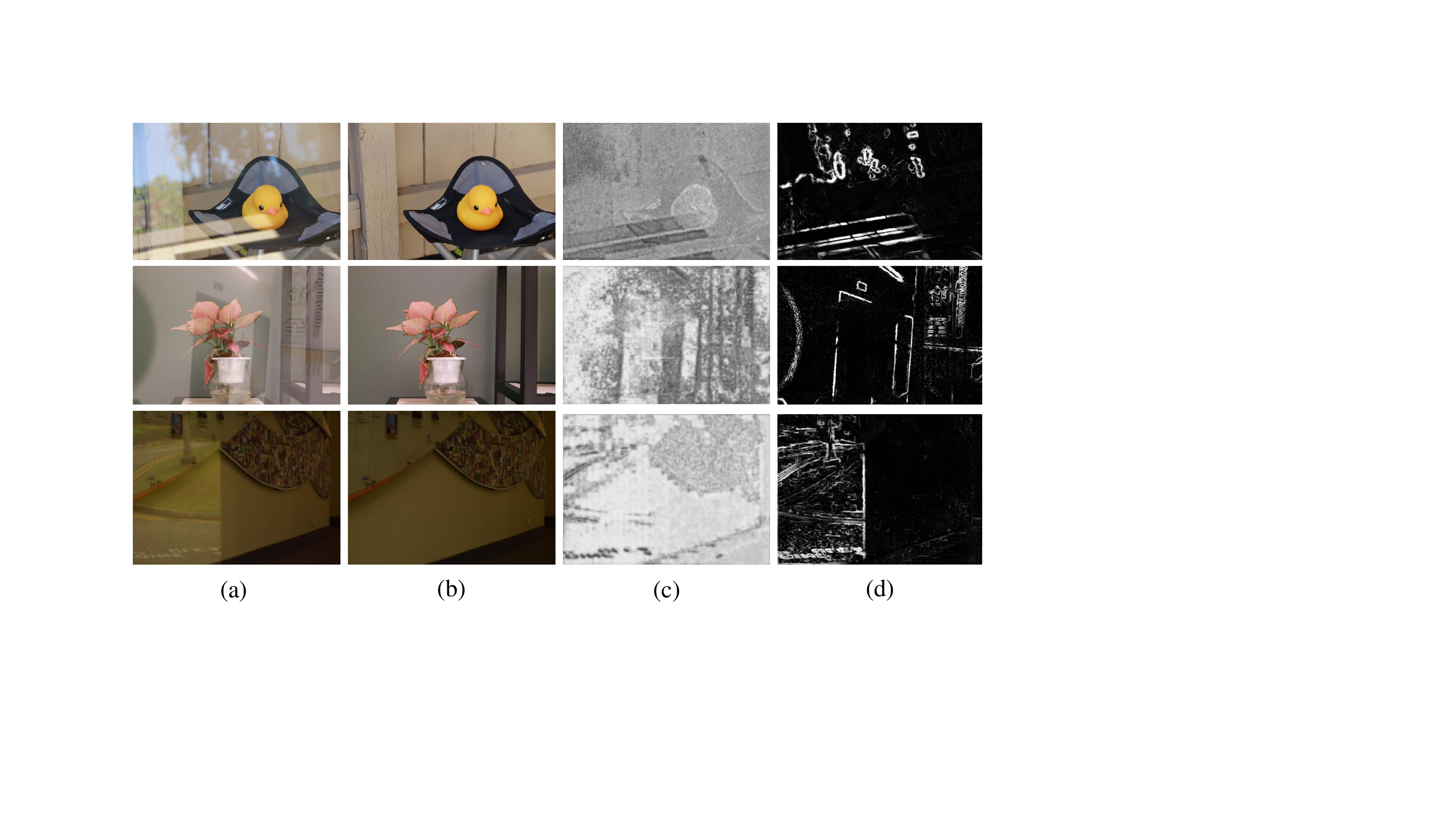}}
    \vspace{-18pt}
\caption {Estimated results of the local reflection locations.  (a) Reflection images; (b) reference images; (c) reflection confidence maps of LANet~\cite{dong2021location}; (d) the estimated location maps of Ours.}

\vspace{-7pt}
\label{fig: location_maps}
\end{figure}

\paragraph{Visualization of the reflection location.}
Figure~\ref{fig: location_maps} shows the estimated location maps of reflection regions, including the previous method (LANet~\cite{dong2021location}) and ours. In their method, LANet employs the linear composition loss to implicitly deduce the reflection confidence maps. Since the distribution of reflection distortions is often uneven, their learned confidence maps could be used to simply depict the locations of the reflection regions. In contrast, We directly utilize the reflection representation obtained from MaxRF as the objective to optimize the reflection detection network (RDNet). Compared to LANet, which represents reflection locations based on weighted composition weights, our approach offers a more explicit manner to directly characterize local reflections.
As a result, the visualization results in Figure~\ref{fig: location_maps} show that our method can estimate the locations of local reflections more accurately and clearly.

\begin{table}[t]
\caption{Ablation study of the components of our proposed SIRR solution. Extra Data indicates that we incorporate the RRW as the additional dataset during the training phase. The PSNR$\uparrow$ metric is used to measure the model performance.}
\vspace{-10pt}

\label{tab:AS_Configurations}
\resizebox{\linewidth}{!}{
\begin{tabular}{cccccc}
\toprule
\multicolumn{3}{c}{Model Configurations} & \multirow{2}{*}{$Nature$} & \multirow{2}{*}{$Real$} & \multirow{2}{*}{$SIR^{2}$} \\ \cmidrule(l{2pt}r{2pt}){1-3} 
RDNet     & RRNet    & Extra Data  &                         &                       &                      \\  \hline

 $\checkmark$         &  $\checkmark$        &                     &     25.37                    &    22.65                   &    24.93                  \\
         &  $\checkmark$        &    $\checkmark$                  &   25.49                      &  23.38                      &  24.82                    \\
  $\checkmark$        &     $\checkmark$     &     $\checkmark$                 & 25.96                   & 23.82                & 25.45   \\ \bottomrule           
\end{tabular}}
\vspace{-10pt}

\end{table}

\begin{table}[t]
\caption{Extension of our proposed innovations to pioneer works. }
\vspace{-10pt}
\centering
\label{tab:AS_extension}
\resizebox{0.9\linewidth}{!}{
\begin{tabular}{lccc}
\toprule
\multicolumn{1}{c}{Model Configurations} & $Nature$                    & $Real$                      & $ SIR^{2} $                         \\ \hline
Zhang \etal~\cite{zhang2018single}                                  & 22.31                     & 20.16                     & 23.07                     \\
\hspace{0.5cm}  +RDNet                                    &    23.02                       &  20.76                         &      23.46                     \\
\hspace{0.5cm}  +RDNet + Extra Data                       &     23.57                      &     21.40                      &  23.79                         \\ \hline
ERRNet~\cite{wei2019single}                               & 22.57                  & 20.67                     & 22.97                     \\
\hspace{0.5cm}  +RDNet                                    &    23.64                       &  21.15                         &      23.57                     \\
\hspace{0.5cm}  +RDNet + Extra Data                       &     24.15                      &     21.73                      &  23.93                         \\ 
\bottomrule
\end{tabular}}
\vspace{-8pt}

\end{table}

\vspace{-15pt}

\paragraph{ Analysis of the components in our framework.}  
Compared with other methods, the core components of our proposed method lie in the reflection-aware guidance network(RDNet) based on MaxRF, and the usage of a real-world dataset collected with the proposed pipeline as additional training data. We conduct experiments to evaluate the impact of essential components of our framework. We compare three models with different configurations:
(i) RDNet + RRNet: the extra data from RRW is not involved in our training dataset, which aims to assess the effect of the extra data in our framework.
(ii) RNet + Extra Data: this aims to assess the effect of the reflection-location guidance in our framework.
(iii) RDNet + RRNet + Extra Data: core components of our framework both are included. This is also our default model configuration.

The quantitative results are reported in Table~\ref{tab:AS_Configurations}. Augmented by the incorporation of supplementary data RRW, our approach exhibits a notable enhancement in performance across various reflection scenarios. This performance improvement is especially noticeable with the $Real$ and $Nature$ datasets. In contrast to the many controlled reflection scenes in $SIR^2$, these two datasets primarily encompass a range of wild reflection scenes. Therefore, powered by these two core innovations, our method can achieve the best de-reflection effects and exhibit superior generalization across real-world reflection scenarios.

\vspace{-8pt}
\paragraph{Extension of our proposed methods.}
The proposed innovations also could be extended to the previous methods. In the ablation study, we apply the reflection location guidance and the additional data expansion for methods~\cite{zhang2018single, wei2019single}. For the former, we cascade the RDNet before the network framework of the previous method. This is in line with our own framework, thereby constructing a two-stage network structure with reflection-aware guidance. As for the latter, we further supplement the cascaded framework by incorporating RRW data as an additional data expansion. The experimental results can be found in Table ~\ref{tab:AS_extension}. This indicates our proposed innovations benefit pioneer methods as well, enhancing the generalization in the real reflection scenarios.


%% file: tabs/performance_comparisons.tex
\begin{table*}[t]
\centering
\caption{Quantitative comparisons on the real reflection benchmarks. The best results are in \textbf{bold}, and the second-best results are \underline{underlined}.
}
\label{tab:performance}
\begin{tabular}{crcccccccc}
\toprule
\multirow{2}{*}{Methods}   & \multirow{2}{*}{Venue} & \multicolumn{2}{c}{$Nature$ (20)} & \multicolumn{2}{c}{$Real$(20)} & \multicolumn{2}{c}{$SIR^{2}$(454)} & \multicolumn{2}{c}{$Average$(494)} \\  \cmidrule(l{2pt}r{2pt}){3-4} \cmidrule(l{2pt}r{2pt}){5-6} \cmidrule(l{2pt}r{2pt}){7-8} \cmidrule(l{2pt}r{2pt}){9-10}
                           &                        & PSNR$\uparrow$         & SSIM$\uparrow$        & PSNR$\uparrow$        & SSIM$\uparrow$       & PSNR$\uparrow$       & SSIM$\uparrow$       & PSNR$\uparrow$         & SSIM$\uparrow$         \\ \hline
Input Image                & -                      & 20.44        & 0.785       & 18.96       & 0.733      & 22.76      & 0.885      & 22.51        & 0.884        \\ \hline
BDN ~\cite{yang2018seeing}                       &  ECCV 2018                      &   18.83            &  0.738            &  18.64           &  0.726          &    21.61        &   0.854          &   21.50           &    0.844          \\
  FRS ~\cite{yang2019fast}                     & CVPR 2019                      &  20.01            &   0.756          &  18.63           &    0.719        &  22.23          &   0.867         &   21.99           &   0.867           \\
Zhang \etal~\cite{johnson2016perceptual} &  CVPR 2018                      &  22.31             &   0.804          &  20.16           &  0.767          &  23.07          &    0.869        &    22.92          &   0.862           \\
ERRNet~\cite{wei2019single}                     &  CVPR 2019                     &  22.57            &    0.807         &   20.67          &    0.781        &     22.97       &   0.885         &            22.85     &   0.877        \\
RMNet~\cite{wen2019single}                     &  CVPR 2019                      &   21.08           &   0.730           &  19.93           &  0.718          &   21.66         &   0.843         &  21.57            &  0.834            \\
Kim \etal~\cite{kim2020single}                      & CVPR 2020                 &   20.10         &    0.759         &    20.22         & 0.735           &   23.57         &    0.877        &      23.30        &    0.886          \\
IBCLN~\cite{li2020single}                      & CVPR 2020                 &   23.90           &    0.787         &    21.42         & 0.769           &   24.05         &    0.888        &      23.94        &    0.878          \\
YTMT~\cite{hu2021trash}                      &  NerIPS 2021                     &   20.69           &    0.777         &   22.94          &   0.815          &  23.57          &  0.889          &    23.43          &   0.882           \\
LANet~\cite{dong2021location}                &   ICCV 2021                     &  23.51            &   \underline{0.810}          &  \underline{23.40}           &    \textbf{0.826}        &    23.04         &  0.898          &    23.07          &  0.891           \\
PNACR~\cite{wang2023personalized}                     & ACM MM 2023                   & \underline{23.93}              & 0.807            &   22.57           &    0.806        &   24.14         & 0.894            &   24.06           &   0.888           \\
DSRNet~\cite{hu2023single}                      &    ICCV 2023                    &     21.24         &  0.789           &   22.32          &   0.806         &  \underline{24.91}           &   \underline{0.902}         &   \underline{24.65}            &    \underline{0.893}          \\
Ours                       & -                      & \textbf{25.96}         & \textbf{0.843}       & \textbf{23.82}       & \underline{0.817}      & \textbf{25.45}      & \textbf{0.910}      &  \textbf{25.40}      & \textbf{0.904}  \\   \bottomrule   
\end{tabular}
\end{table*}

%% file: Sec5_Conclusion.tex
\begin{figure}[t]
\captionsetup{type=figure}
\centering
\resizebox{\linewidth}{!}{
\includegraphics[width=1.0\textwidth]{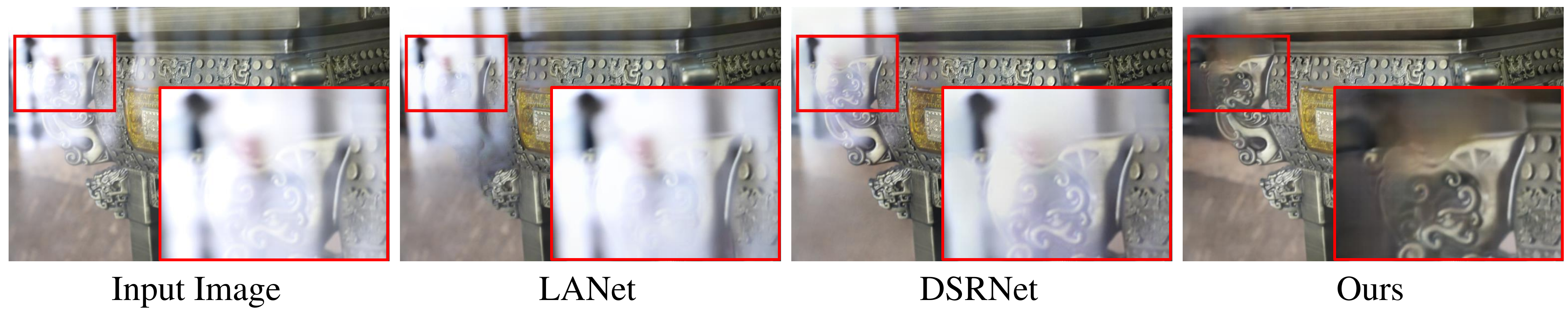}}
    \vspace{-20pt}
\caption {Failure case in the saturated reflection scene. }
\vspace{-15pt}
\label{fig: limitation}
\end{figure}

\section{Conlusion}
\vspace{-5pt}
In this paper, we revisit the physical formulation of the reflection degradation imaging. This motivates us to propose a more applicable pipeline for real-world reflection data collection. Our pipeline is conducted in the video collection mode, enabling us to collect a large-scale, high-quality reflection training dataset at a lower cost, named RRW.  Furthermore, we propose MaxRF, which explicitly identifies the locations of reflection distortions from aligned reflection image pairs, and develop a location-aware guidance framework for SIRR. With the support of our customized framework and RRW, our solution achieves superior performance against SOTA methods in real-world benchmarks. Meanwhile, we also suggest that these innovations could enhance the reflection removal performance of previous methods, hoping to inspire future research in this field.





\vspace{-10pt}
\paragraph{Limitations.} Our method may encounter challenges in scenarios with saturated reflections. 
The intensity of these saturated reflections is extremely high, making the underlying transmission contents nearly invisible in these saturated regions.  
Figure~\ref{fig: limitation} illustrates this limitation. Although our method can mitigate these reflections to a certain degree, residual artifacts may remain.
Given that the transmission contents loss occurs in the saturated regions, which often constitute a  portion of the image, we intend to integrate semantic information to further enhance the restoration performance in future work.

